\documentclass{article}
\usepackage{ijcai21}

\usepackage{times}
\usepackage{soul}
\usepackage{url}
\usepackage{paralist}
\usepackage[hidelinks]{hyperref}
\usepackage[utf8]{inputenc}
\usepackage[small]{caption}
\usepackage{graphicx}
\usepackage{amsmath,amsfonts}
\usepackage{amsthm}
\usepackage{booktabs}
\usepackage{algorithm}
\usepackage{algorithmic}
\usepackage{amsmath, amssymb}
\usepackage{mathtools}
\newcommand{\defeq}{\vcentcolon=}
\usepackage{multirow}
\usepackage{arydshln}

\usepackage{subfigure}
\newtheorem{theorem}{Theorem}
\usepackage{color}
\newcommand{\blue}[1]{{\color{blue}{#1}}}

\usepackage{CJKutf8}

\usepackage[switch]{lineno} 
\newcommand*\patchAmsMathEnvironmentForLineno[1]{%
  \expandafter\let\csname old#1\expandafter\endcsname\csname #1\endcsname
  \expandafter\let\csname oldend#1\expandafter\endcsname\csname end#1\endcsname
  \renewenvironment{#1}%
     {\linenomath\csname old#1\endcsname}%
     {\csname oldend#1\endcsname\endlinenomath}}% 
\newcommand*\patchBothAmsMathEnvironmentsForLineno[1]{%
  \patchAmsMathEnvironmentForLineno{#1}%
  \patchAmsMathEnvironmentForLineno{#1*}}%
\AtBeginDocument{%
\patchBothAmsMathEnvironmentsForLineno{equation}%
\patchBothAmsMathEnvironmentsForLineno{align}%
\patchBothAmsMathEnvironmentsForLineno{flalign}%
\patchBothAmsMathEnvironmentsForLineno{alignat}%
\patchBothAmsMathEnvironmentsForLineno{gather}%
\patchBothAmsMathEnvironmentsForLineno{multline}%
}

\def \x {\mathbf{x}}
\def \y {\mathbf{y}}
\def \z {\mathbf{z}}

\def \f {\mathbf{f}}
\def \h {\mathbf{h}}
\def \a {\mathbf{a}}

\def \CE {\mathcal{E}}

\def \g {\mathbf{g}}

\def \u {\mathbf{u}}
\def \v {\mathbf{v}}

\def \tx {\mathbf{\tilde{x}}}
\def \MH {{\rm\bf MH}}

\def \LN {{\rm LN}}
\def \FFN {{\rm FFN}}
\def \MLP  {{\rm\bf MLP}}

\def \ty {{\tilde{\bf y}}}
\def \th {{\tilde{\bf h}}}

\def \M {\mathbf{M}}
\def \B {\mathbf{B}}

\def \P {\mathbf{P}}

\def \I {\mathbf{I}}

\def \A {\mathbf{A}}

\def \mT {\mathcal{T}}
\def \mC {\mathcal{C}}
\def \WE {\mbox{WE}}
\def \PE {\mbox{PE}}

\def \R {\mathbb{R}}

\def \bmu {\pmb{\mu}}
\def \balpha {\pmb{\alpha}}

\def \bpsi {\pmb{\psi}}
\def \bsigma {\pmb{\sigma}}

\def \btheta {\pmb{\theta}}

\title{Progressive Open-Domain Response Generation with Multiple Controllable Attributes}

\author{
Haiqin Yang$^1$\and
Xiaoyuan Yao$^1$\and
Yiqun Duan$^{1}$\and
Jianping Shen$^{1}$\and
Jie Zhong$^{1}$\And
Kun Zhang$^2$\\
\affiliations
$^1$Ping An Life Insurance Company of China\\
$^2$Carnegie Mellon University\\ % , Pittsburgh, PA, USA
\emails
hqyang@ieee.org, yaoxiaoyuan617@pingan.com.cn, duanyiquncc@gmail.com, jpshen2008@qq.com, zhongjie@pingan.com.cn, kunz1@cmu.edu
}

\begin{document}
%\linenumbers 

\maketitle

\begin{abstract} %  by generating open-domain response with multiple controllable attributes In social chatbots, it is desirable to 
It is desirable to include more controllable attributes to enhance the diversity of generated responses in open-domain dialogue systems.  However, existing methods can generate responses with only one controllable attribute or lack a flexible way to generate them with multiple controllable attributes.  In this paper, we propose a Progressively trained Hierarchical Encoder-Decoder (PHED) to tackle this task.  More specifically, PHED deploys Conditional Variational AutoEncoder (CVAE) on Transformer to include one aspect of attributes at one stage.  A vital characteristic of the CVAE is to separate the latent variables at each stage into two types: a global variable capturing the common semantic features and a specific variable absorbing the attribute information at that stage.  PHED then couples the CVAE latent variables with the Transformer encoder and is trained by minimizing a newly derived ELBO and controlled losses to produce the next stage's input and produce responses as required.  Finally, we conduct extensive evaluations to show that PHED significantly outperforms the state-of-the-art neural generation models and produces more diverse responses as expected.  
\end{abstract}

\section{Introduction}
Developing human-like conversational agents is a long-lasting goal of artificial intelligence.  Recently, thanks to the availability of a plethora of conversation data on the Internet and the booming of deep learning technologies, researchers have been attracted to explore end-to-end data-driven approaches to building social chatbots~\cite{DBLP:journals/tois/HuangZG20}. % in both academia and industry ,DBLP:journals/ftir/GaoGL19

\begin{figure}[ht]
\small
\centering%\footnotesize
\begin{tabular}%{@{~}l@{~}l@{~}l@{~}}
{@{~}l@{~}p{7.4cm}%p{7.5cm}
}
\hline % & 
%& 全场比赛结束。乌迪内斯2-1米兰。%& 
%\\ 
{\multirow{1}{*}{Post}} & The match is over.  Udinese 2-1 Milan. \\\hdashline
%R$_{\rm \blue{\bf H}}$ & {\blue{\bf 恭喜}乌迪内斯2:1米兰，新赛季继续加油！} % &\\ & \textbf{\blue{Congratulations}} Udinese 2:1 Milan, keep going in the new season!  \\ \hline %\blue{\bf Congratulations} Udinese 2:1 Milan, keep going in the new season!
R$_{{\rm {H}}}$ %& \textbf{\blue{恭喜}}乌迪内斯2:1米兰，新赛季继续加油！\\
& $\overleftrightarrow{\mbox{Congratulations}}$ Udinese, keep going in the new season!
%\blue{\bf 恭喜}乌迪内斯2:1米兰，新赛季继续加油！\\ & \textbf{\blue{Congratulations}} Udinese 2:1 Milan, keep going in the new season!
\\ %\hdashline %\hline
R$_{\rm HI}$ %& \textbf{\blue{ 哈哈}}，\underline{米兰这是要逆天吗？} \\
&  $\overleftrightarrow{\mbox{Haha}}$, \underline{is} Udinese going to offence God's will\underline{?} \\%\hdashline %\hline
R$_{\rm HIL}$ %& \textbf{\blue{哈哈}}，不知道\underline{\blue{为什么}}这个时候还是\textbf{\blue{很开心}} \\  
&  $\overleftrightarrow{\mbox{Haha}}$, don't know \underline{{why}} this time I am still $\overleftrightarrow{\mbox{very happy}}$  \\
\hline
\end{tabular}
\caption{An example of generated responses with progressively fed attributes:  The word with $\leftrightarrow$ on top indicates its high specificity to the Happy emotion.  The \underline{underlined word} denotes the \underline{I}nterrogative tone while the third character of ${\rm L}$ requires generating a long response.\label{fig:ex_illustration}} % in the subscript DBLP:conf/naacl/SordoniGABJMNGD15,
\end{figure}
Nowadays, sequence-to-sequence (Seq2Seq) models~\cite{DBLP:conf/aaai/SerbanSBCP16} have been adopted to generate conversations due to their scalability and promising capability in capturing language-independence to implicitly learn semantic and syntactic relations between message-response pairs and contextual dependencies.  However, they usually tend to generate ``safe responses'', such as ``I do not know'' and ``OK'', because the vanilla Seq2Seq models are prone to only memorize high-frequency responses in the data~\cite{DBLP:conf/aaai/SerbanSLCPCB17,DBLP:conf/aaai/XingWWHZ18}.  Various neural generation methods have been proposed to incorporate different controllable attributes or rich information into the Seq2Seq framework to enhance the generation diversity.  The attributes may include length~\cite{DBLP:conf/emnlp/KikuchiNSTO16}, sentiment and emotion~\cite{DBLP:conf/icml/HuYLSX17,DBLP:conf/acl/WangZ18,DBLP:conf/aaai/ZhouHZZL18,DBLP:conf/acl/RashkinSLB19}, tone~\cite{DBLP:conf/acl/HuangKGx18,DBLP:conf/acl/BiGLS19}, specificity~\cite{DBLP:conf/acl/ChengXGLZF18,DBLP:conf/naacl/SeeRKW19}, and meta-words~\cite{DBLP:conf/acl/XuWTHSW19}.  Recently, it is desirable to generate responses with multiple controllable attributes because it can allow social chatbots to create more human-like responses and manifest more intelligence from different angles~\cite{DBLP:journals/tois/HuangZG20,DBLP:conf/aaai/ZhengZHM20}.  However, existing methods usually generate responses with only one controllable attribute or fail to provide a flexible way to generate them with multiple controllable attributes~\cite{DBLP:conf/naacl/SeeRKW19}.

% commonsense knowledge,DBLP:conf/naacl/KoDL19 bases~\cite{DBLP:conf/aaai/GhazvininejadBC18,DBLP:conf/ijcai/ZhouYHZXZ18,DBLP:conf/aaai/YoungCCZBH18}, and , and , topics~\cite{DBLP:conf/aaai/XingWWLHZM17},.  Additional information includes the content~\cite{DBLP:conf/coling/MouSYL0J16,DBLP:conf/emnlp/YaoZFZY17} ,DBLP:conf/naacl/LiGBGD16,DBLP:conf/nips/ZhangGGGLBD18 style~\cite{DBLP:conf/emnlp/WangJBN17,DBLP:conf/sigdial/OrabyRTSLW18}
%extensions, e.g., control specificity and hierarchical structure, have been proposed to enhance the diversity of the generated responses~\cite{DBLP:conf/naacl/LiGBGD16,DBLP:conf/emnlp/ZhangZ19}.  It is still hard to efficiently generate diverse responses, especially with multiple controlled attributes, which has been a challenging task in developing personalized dialogues~\cite{DBLP:journals/tois/HuangZG20,DBLP:conf/aaai/ZhengZHM20}.  In this paper, we aim to generate open-domain responses with multiple controllable attributes, an under-explored research problem. 

In this paper, we develop a new framework, the Progressively trained Hierarchical Encoder-Decoder (PHED), to tackle this task.  As illustrated in Fig.~\ref{fig:ex_illustration}, PHED effectively generates responses with three different aspects of controllable attributes in a progressive way: the Happy emotion in the first aspect, the Interrogative tone in the second one, and the long response generation requirement in the third one.  PHED enjoys prominent properties: (1) It acts as an interface for developers to customize responses by tailoring the attributes partially or fully.  In~\cite{DBLP:conf/acl/XuWTHSW19}, all controllable attributes need to be preset.  Differently, our PHED can output each stage of responses with one desired attribute at one stage. (2) The framework is extensible and scalable.  More aspects of attributes can be easily incorporated in the generation procedure.  This is different from existing work on text generation with multiple attributes~\cite{DBLP:conf/nips/LogeswaranLB18,DBLP:conf/iclr/LampleSSDRB19,DBLP:conf/emnlp/ShaoHWXZ19}.

%enjoys three advantages: (1) the generation process is controllable and flexible.  The attributes act as an interface that allows developers to customize responses by tailoring the set of attributes.  Our setting is also different from~\cite{DBLP:conf/acl/XuWTHSW19}, which can only generate responses with all preset attributes.  Our model can generate responses with partially controlled attributes or all controlled attributes simultaneously; (2) the generation model is explainable as the attributes inform the model, developers, and even end users what kinds of characteristics of responses they expect before the responses are generated; (3) the generation is widely applicable and scalable.  By taking users' emotions~\cite{DBLP:conf/aaai/ZhouHZZL18}, sentence functions~\cite{DBLP:conf/acl/HuangKGx18, DBLP:conf/acl/BiGLS19}, and response length~\cite{DBLP:conf/emnlp/KikuchiNSTO16} as considered attributes, we can address the generation issue in a unified way.  Especially, we can easily extend the architecture progressively to incorporate more attributes.  This setting also distinguishes our work from text transfer with multiple attributes~\cite{DBLP:conf/nips/LogeswaranLB18,DBLP:conf/iclr/LampleSSDRB19}. DBLP:conf/nips/SutskeverVL14,DBLP:journals/corr/BahdanauCB15  and machine translation~\cite{DBLP:journals/corr/BahdanauCB15}

To ensure the relevance of a response to the message and fidelity of the response to the controlled attributes, PHED designs subtle losses under rigorous mathematical derivation.  Specifically, we utilize Transformer because it facilitates the self-attention mechanism for many NLP applications~\cite{DBLP:conf/nips/VaswaniSPUJGKP17}.  To ensure the diversity of the generated responses with controllable attributes, we apply Conditional Variational AutoEncoder (CVAE) and separate the CVAE latent variables into two meaningful types of variables: a joint latent variable capturing semantic features shared among all data and specific latent variables, each of which controls the attribute at the corresponding stage.  The learned CVAE latent variables are then coupled with the encoding information learned at previous stages to {\em explicitly} promote the effect of the specific attributes in generating responses.  Here, we borrow the idea of story completion in~\cite{DBLP:conf/ijcai/Wang019b} to utilize the proved effective architecture of Transformer-based CVAE (T-CVAE) to implement the coupling procedure.  Different from T-CVAE, PHED does not share the parameters in the encoder and the decoder, but contains more CVAE latent variables, which are optimized by a newly derived evidence lower bound (ELBO) and controlled losses.  We conduct extensive evaluations and demonstrate that PHED can generate more diverse responses.%significantly outperforms the state-of-the-art neural generation methods.

%We test our proposed PHED on a large-scale open-domain conversation dataset from Weibo and compare PHED with several state-of-the-art neural generation methods in terms of response relevance, response diversity, accuracy of one-to-many modeling, accuracy of attributes expression, and human judgment.  Evaluation results show that PHED significantly outperforms the state-of-the-art neural generation methods over most of the metrics.  Moreover, in training PHED, we are also inspired by a proved humans' effective learning procedure~\cite{journals/PS/SpieringA08}, i.e., learning in the order from most to least difficult tasks, a similar progressive learning procedure in~\cite{DBLP:conf/iclr/0007MGW20}.mpare PHED with several state-of-the-art neural generation models

The contribution of our work is threefold: (1) a first work to generate diverse responses with multiple controllable nesting attributes; (2) a unified framework to include only one aspect of controllable attributes at one stage, relying on a hierarchical structure that enjoys flexibility and extensibility with rigorous theoretical guarantee; (3) empirical evaluations clearly demonstrating the effectiveness of PHED. % on both automatic and human metrics 

\begin{figure*}[!hbt]
\centering
\subfigure[PHED]{\includegraphics[width=0.6\textwidth]{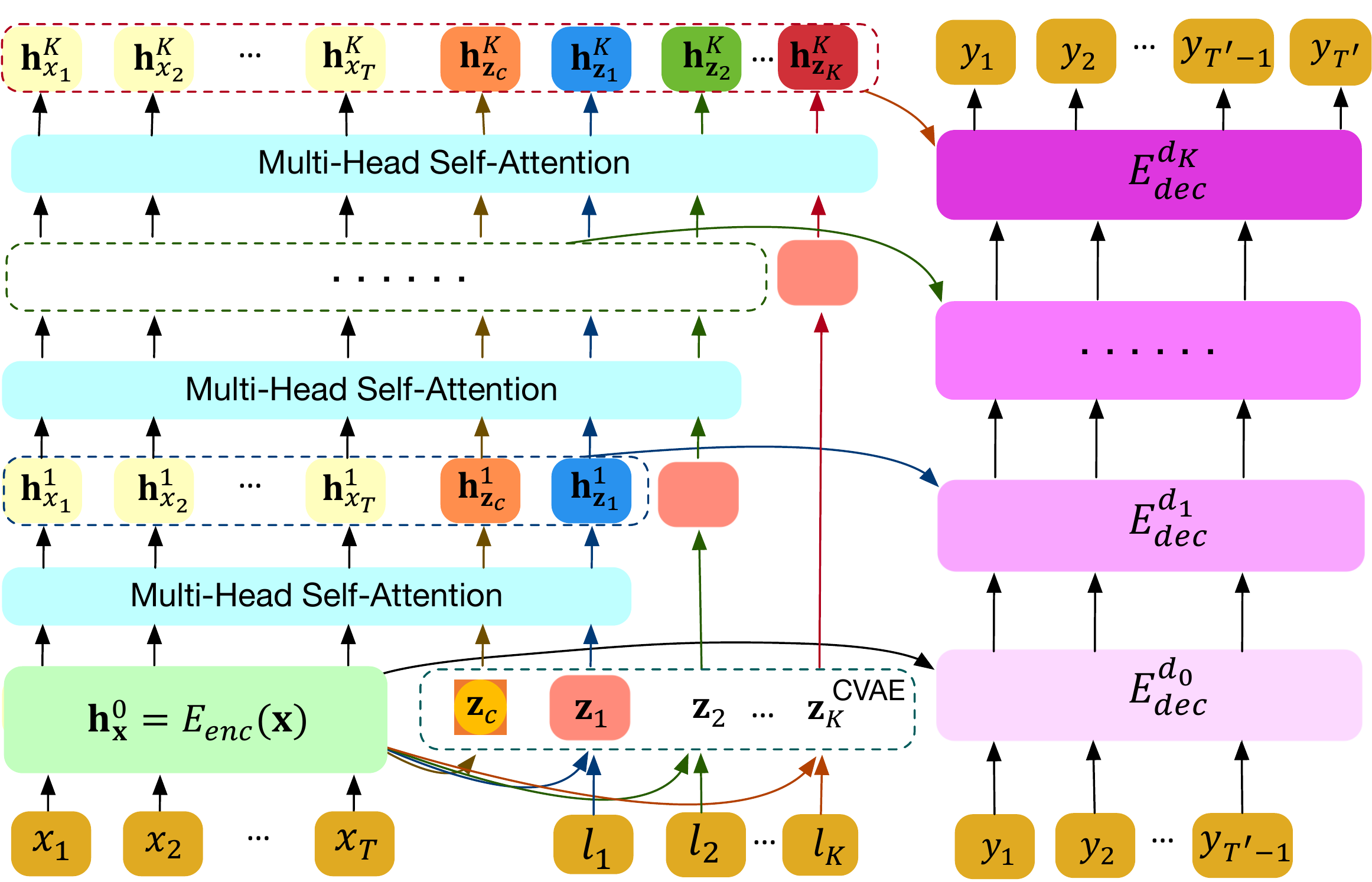}\label{fig:PHED}}%\qquadPHED_Inference
\subfigure[CVAE]{
\includegraphics[width=0.4\textwidth]{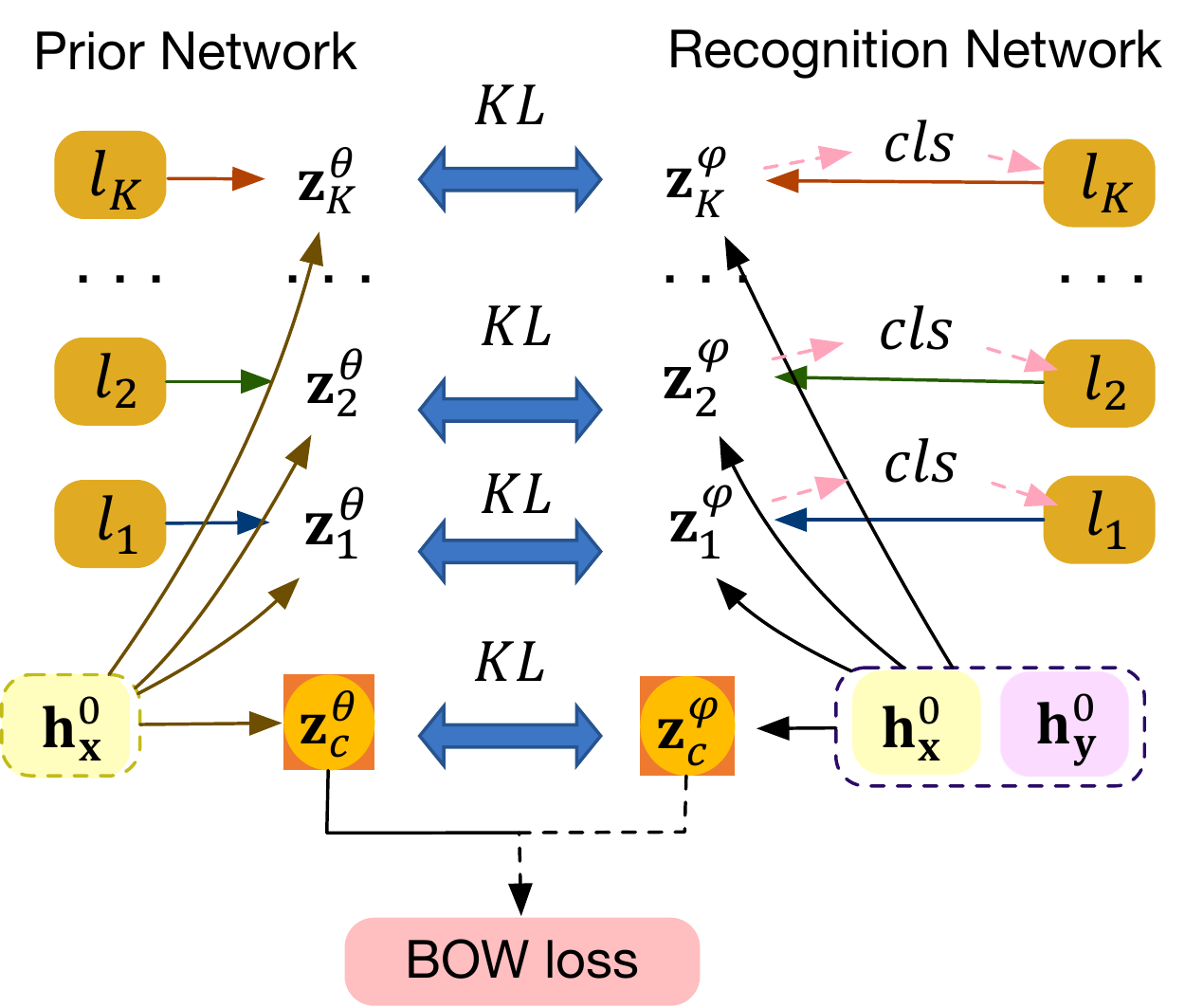}\label{fig:VAE}
}%$\h_\x$ and $\h_\y$
\caption{The architecture of the PHED framework:  Different colors on different variables distinguish the effect of attributes at the corresponding stage.  For the variables and the training procedure, please see the main text.  \label{fig:PHED_desc}%  $\h_\x^0$ and $\h^0_\y$ are defined in Eq.~(\ref{eq:h0})
}
\end{figure*}

\section{Our Proposal}
%We define the task and present our proposed PHED with some essential theoretical results  and its training procedure. 
We present PHED with the theoretical results and its training procedure. 
%\subsection{Task Definition} %  and Preliminaries \label{sec:task_definition}
%Our task is defined as follows: 
%{\bf Preliminaries.} 
\subsection{Preliminaries}
Given a corpus, ${\cal D}=\{(\x_i, c_i, \y_i)\}_{i=1}^N$, where $N$ is the number of message-response pairs, $\x_i=x_{i1}\,x_{i2}\,\ldots\,x_{i|{\x_i}|}$ is a message with $|{\x_i}|$ characters or words, $c_i$ denotes the associated attributes on the response of $\y_i=y_{i1}\,y_{i2}\,\ldots\,y_{i|{\y_i}|}$, the objective is to learn the conditional probability $p(\y|\x, c)$ from the corpus.  Here, the attribute $c=l_1, \ldots, l_K$ enforces the attribute $l_i$  at the $i$-th stage from $K$ pre-defined aspects, e.g., the emotion of happy or sad~\cite{DBLP:conf/naacl/JiaoYKL19}, and the tone of Declarative, Interrogative, or Imperative~\cite{DBLP:conf/acl/HuangKGx18}.  After obtaining $p(\y|\x, c)$, given a message $\x$ and a specific attribute $c$, we will generate response $\y$ accordingly.

% In this work, we assume that $c$ is given as input for response generation.  Later, we will describe how to obtain $c$ with X.  %

%\subsection{Progressively Trained Hierarchical Encoder-Decoder (PHED) Framework} \label{sec:PHED}
%Since the controllable attributes $c$ come from different domains, a simultaneous learning mechanism may yield negative influence~\cite{DBLP:conf/naacl/SeeRKW19}.  Though the response $\y$ has included multiple attributes, we enforce only one attribute at one stage and train the model progressively.  In our framework, $c=l_1, \ldots, l_K$, where $l_i$ denotes the specific controllable attribute enforced at the $i$-th stage.

We propose Progressively trained Hierarchical Encoder-Decoder (PHED), shown in Fig.~\ref{fig:PHED}, to enforcing controlling only one aspect of attributes from the data at one stage.  The basic structure of PHED resembles T-CVAE~\cite{DBLP:conf/ijcai/Wang019b}, but distinguishes the CVAE variables to $\z_c\in\R^{d_{\z_c}}$ for capturing common semantic features and $\z_i\in\R^{d_{\z_i}}$ for capturing individual features at the $i$-th stage, where $d_{\z_c}$ and $d_{\z_i}$ denote the size of $\z_c$ and $\z_i$, respectively. 

To relieve the burden of the model expression, we define the vanilla Transformer layer for the encoder ($\mT_e$) and the decoder ($\mT_d$) as follows: 
\begin{align}\label{eq:Tencoder}
     \h^i&=\mT_{e}(\h^{i-1})\defeq\left\{
    \begin{array}{@{}l@{}l@{}}
    \A &=\MH_{\CE_e}(\h^{i-1}, \h^{i-1}, \h^{i-1}), \\
    \B &= \LN(\h^{i-1}+\A),\\
    \h^i &= \LN(\FFN(\B)+\B),
    \end{array}
    \right.\\\label{eq:Tdecoder}
    \h_d^i&=\mT_{d}(\h_d^{i-1}, \h_e) :=\left\{
    \begin{array}{@{}l@{}l@{}}
    %\h_s &= \mT_{e}(\h_d^{i-1}, \h_d^{i-1}, \h_e, \h_e) \\
    \f &=\MH_{\CE_d}(\h_d^{i-1}, \h_e, \h_e), \\
    \g &= \LN(\h_d^{i-1}+\f),\\
    \h_d^i &= \mT_{e}(\g).
    %\h_d^i &= \mT_{e}(\h_s, \h_s, \h_s, \h_s).
    %\A &=\MH_{\CW_{d1}}(\e, [\v; \e], [\v; \e]), \\
    %\B &= \LN(\e+\A),\\
    %\e_o &= \LN(\FFN_{\CW_{d2}}(\B)+\B),
    \end{array}
    \right.
\end{align}%  Here, $H$ is the size of hidden features in Transformer. 
Here, all the hidden size in the Transformer layer is $H$.  $\h^i, \h_d^i\in\R^H$ denotes the output of the encoder and the decoder at the $i$-th Transformer layer, respectively.  $\MH_{\CE_e}$ and $\MH_{\CE_d}$ is the multi-head attention network with the input of query, key, and value in the encoder and the decoder, respectively.  $\LN$ denotes the operation of layer normalization and $\FFN$ is a feed forward neural network.   %For simplification, we usually let $\q, \k, \v\in\R^d_k$. $\in\R^H$ the output of the decoder at the $i$-th sub-layer, and $\h_e\in\R^H$ in $\mT_d$ an hidden feature from the encoder.  The input, $\q, \k, \v\in\R^H$ in $\MH_{\CE_e}$, corresponds to the query, key, and value, respectively, of Their inputs are , respectively. 

Borrowing T-CVAE in~\cite{DBLP:conf/ijcai/Wang019b}, we define: %\\\nonumber
\begin{equation}\label{eq:CVAE_nets}
\z=\mC(\Psi, \a) \defeq \left\{\begin{array}{@{\!}r@{~}l}
\mbox{I.} &
\v = {\MH}_\Psi(\balpha, \a, \a),\\
\mbox{II.} &
\left[ \begin{array}{@{~}c@{~}}
\bmu \\
\log\left(\bsigma^2\right) 
\end{array}
\right ]\! = \! {\MLP}(\v),\\\mbox{III.} & \z \sim N(\bmu, \bsigma^2\I),
\end{array}
\right.\!
\end{equation}
Hence, $\z$ is sampled from an isotropic Gaussian distribution with mean ($\bmu$) and variance ($\bsigma$) computed in two steps.  In Step I, a hidden feature $\v$ is computed from a multi-head attention network on $\Psi$, ${\MH}_\Psi$, which takes three inputs, i.e., $\balpha$ (a random initialized context vector) for the query, and $\a$ for the key and the value, respectively.  In Step II, $\v$ is fed  to a multi-layer perceptron ($\MLP$) to determine the mean and the variance simultaneously.

\subsection{Model and Theory}
Let $\h^i\in\R^H$ and $E_{dec}^{d_i}\in\R^H$ be the output of the encoder and the decoder at the $i$-th stage, respectively.  $d_i$ is the number of layers in the decoder up to $i$-th stage, $i=1,\ldots, K$.  When $i=0$, we compute $\h^0_{\x}$ and $\h^0_{\y}$ by the first Transformer layer: 
\begin{align}\label{eq:h0}
&\h^0=\h^0_{\x}=E_{enc}(\x)\defeq \mT_{e}(\tx), \quad
\h_\y^0=E_{enc}(\y), %\defeq\mT_{e}(\ty, \ty, \ty, \ty). , \tx, \tx, \tx
\end{align} 
where $\tx$ is the sum of token embedding and position embedding on $\x$, i.e., $\WE(\x)+\PE(\x)$.%; likewise $\ty$ for $\y$.

The decoder of the first Transformer layer is computed by  
\begin{align}
   E_{dec}^{d_0} =  \mT_{d}(\ty, \h_\x^0),\mbox{ where } \ty=\WE(\y)+\PE(\y).% , \h_\y^0
\end{align}

The corresponding CVAE variables at the $i$-th stage of the {\em recognition network} and the {\em prior network} are:
\begin{align}\label{eq:regnet_c}
& \z_{c}^i(\bpsi)=\mC(q_{\bpsi, c}^i, [\h_\x^0; \h^0_\y]),  \quad \z_i(\bpsi)=\mC(q_{\bpsi}^i, [\h_\x^0; \h^0_\y]), \\
%\end{align}
%\end{equation}
%and the  by 
%\begin{equation}
%\begin{align}
\label{eq:priornet_i}
&\z_c^i(\btheta)=\mC(p_{\btheta, c}^i, \h_\x^0), \quad \z_i(\btheta)= \mC(p_{\btheta}^i, \h_\x^0).
\end{align}
Obviously, the difference between the two networks lies in whether the multi-head attention network attends to the decoder $\h^0_\y$ or not.  It is noted that $\h_\x^0$ and $\h^0_\y$, rather than $\h^{i-1}$, are applied to learn the parameters of both networks because $\h^{i-1}$ has absorbed the attribute information in all previous stages and may contaminate the original data. 

\if 0
\begin{algorithm}[ht!]
\caption{Online Inference.} 
\label{alg:Framwork} 
\begin{algorithmic}[1] 
\Require 
The input utterance $\u$; the mention-concept set $\M=\bigcup_{r=1}^4 \M_{r}$, where $r\in\{\Action, \Question, \Argument, \Problem\}$, $\M_{r}=\{(m_{r, m, n}, c_{r, m}), n\in [1, N_m], m\in [1, M]\}$, $M$ is the number of concepts within the intent-role $r$, $N_m$ is the number of mention-concept pairs in $c_{r, m}$; $f$ is a phrase embedding function; $\delta$ is a parameter to filter out dissimilar mentions; $K$ is the number of nearest neighbors; the pattern set $\P$.
%$f$ is a phrase embedding function; %$K$ is the number of nearest neighbors;
\Ensure The set of intent-role mentions with concept ID, $Result$;
\end{algorithmic} 
\end{algorithm}
\fi 
%to update $\z_c^i$ and yield the newly introduced latent variable $\z_i$.

We train PHED in the following progressive way: 
\begin{compactenum}%\if 0We first attain the CVAE variables and derive $\h^1$ and $E_{dec}^{d_1}$ afterwards. 
\item The CVAE variables $\z_c^1$ and $\z_1$ are sampled from the {\em recognition network} defined in Eq.~(\ref{eq:regnet_c}) by setting $i=1$.  Next, $\h^1$ and $E_{dec}^{d_1}$ are then computed by the newly stacked Transformer layer on the concatenation of $\h^0$, $\z_c^1$, and $\z_1$ (i.e., $\th^1=[\h^0; \z_c^1; \z_1]$): 
\begin{align}\label{eq:h1}
\!\!\h^1=\mT_{e}(\th^1),\quad E_{dec}^{d_1}=\mT_{d}(E_{dec}^{d_0}, \th^1),
\end{align}
We highlight two remarks: (1) The effect of the CVAE variables $\z_c$ and $\z_1$ is realized by the multi-head self-attention on $\th^1$. (2) The input of $\mT_d$ is slightly different from the standard Transformer.  That is, the self-attention in PHED is applied at the same stage, not from scratch.  It can then enhance the impact of the CVAE variables at the corresponding stage.

%, rather than connecting the output of the encoder to the initial input of the decoder. %~\cite{DBLP:conf/nips/VaswaniSPUJGKP17}. 
% It is worth noting that different from standard Transformer conducting self-attention on the output of the encoder and each Transformer block in the decoder, we only place self-attention on the corresponding layers of the encoder and the decoder at the same stage.
%$E_{dec}^{d_1}(\y) = \DEC(\h^1, \y)$.  %\fi 
\item At the $i$-th stage ($i\ge 2$), we fix the parameters learned at previous stages and sample $\z_c^i$ and $\z_i$ from the {\em recognition network} defined in Eq.~(\ref{eq:regnet_c}) and compute $\h^i$ and $E_{dec}^{d_i}$ by a newly stacked Transformer layer: % are computed 
\begin{align}\label{eq:h_i}
\!\!\h^i=\mT_{e}(\th^i), \quad  E_{dec}^{d_i}=\mT_{d}(E_{dec}^{d_{i-1}}, \th^i), 
\end{align}
where $\th^i=[\h^{i-1}; \z_i]$.  Note that $\th^i$ does not include $\z_c^i$ because it has been absorbed in $\h^{i-1}$. 
%We emphasize that the network parameters for $\z_c^i$, $\z_i$, $\mT_e$, and $\mT_d$ are updated at the current stage by fixing the learned parameters at previous stages. 
%\end{equation}  
%It is noted that since $\h^{i-1}$ has absorbed the attribute information in all previous stages, we only apply $\h_\x^0$ and $\h^0_\y$ to update $\z_c^i$ and yield the newly introduced latent variable $\z_i$.  
% , \th^i, \th^i, \th^i
% with the concatenation of $\h^{i-1}$ and $\z_i$ as the initial input, i.e., the initial input of $\q$, $\k$, and $\v$  is $[\h^{i-1}, \z_i]$, .  Similarly, $E_{dec}^{d_i}$ is obtained by the decoder of a newly stacked Transformer block with the initial query from $\h^i$ and the initial key and value from $E_{dec}^{d_{i-1}}$.  
\item Step 2 continues until we reach the $K$-th stage.  The parameters are learned by the {\bf Multi-stage Training} procedure detained in Sec.~\ref{sec:training}.
\end{compactenum}

By the above generation mechanism, we can derive the following theorem to compute the conditional probability:
\begin{theorem}
Given the above defined notations, the conditional generation probability can be computed by
\begin{align}\nonumber
&p(\y|\x,c=l_1\ldots l_K)\\\label{eq:p_theta_final}
=& p(\y|\h^K, l_K)\cdot \prod_{k=1}^K p(\h^{k}|\h^{k-1}, l_k) \cdot p(\h^0|\x),
\end{align}
and the evidence lower bound at the $i$-th ($i\ge 1$) stage is
\begin{align} \label{eq:MLEELBO} 
%\log p_{\btheta}^i(\y|\x, l_1...l_i)\geq & L_{ELBO}  %\\ %}_{Q} \underbrace_{L_{ELBO}}
&\log p_{\btheta}^i(\y|\x, l_1...l_i)\! \geq\!\underbrace{-L_{KL}^{c, i}\! -\! L_{KL}^i\! -\! L_{M}^i}_{L_{ELBO}},\mbox{~where}
%\end{align}
%where 
%\begin{align}  
\\%\nonumber %\label{eq:KL_c}
& L_{KL}^{c, i} \defeq D_{KL}(q_{\bpsi, c}^i(\z_{c}^i|\h^0, \y)\parallel p_{{\btheta}, c}^i(\z_{c}^i|\h^0)), \\%\nonumber %\label{eq:KL_i}
  &  L_{KL}^i \defeq D_{KL}(q_{\bpsi}^i(\z_{i}|\h^0, \y, l_i)\parallel p_{{\btheta}}^i(\z_{i}|\h^0, l_i)), \\%\nonumber %\label{eq:MLE}
   & L_{M}^i \defeq -E_{\z_c\sim q_{\bpsi, c}^i,  \z_{i} \sim q_{\bpsi}^i}[\log p_{{\btheta}}^i(\y|\h^{i-1}, l_i)].
% & L_{M}^i\!\! = \!\!-E_{\z_c\sim q_{\bpsi, c}^i,  \z_{i} \sim q_{\bpsi}^i}[\log p(\y|\h^{i-1}, \tilde{\z}_i, l_i)].
\end{align}
\end{theorem}
%\noindent 
The proof is provided in the Appendix.  Eq.~(\ref{eq:p_theta_final}) holds due to the variable dependency and the Markov chain on $\h^i$.  Here, we only consider the Markov property and leave the variants of including more hidden states as a future work.  Note that in Eq.~(\ref{eq:MLEELBO}), the derived ELBO ($L_{ELBO}$) consists of not only the expected log-likelihood estimator, $L_{M}^i$, but also two separated KL divergences, $L_{KL}^{c, i}$ and $L_{KL}^i$ to control $\z_c^i$ and $\z_c$.  

%see detailed derivation in Appendix~\ref{ap:gp}.  

% we will concatenate $\z_c^1$ and $\z_1$ with $\h^0$, i.e., $[\h^0; \z_c^1; \z_1]$, as the initial input of the encoder of the modified Transformer block at the first stage while $[\h^{i-1}, \z_i]$ ($i>1$) for other stages.
\subsection{Losses and Training}\label{sec:training}
%{\bf Losses and Training.}  
Other than the derived KL divergence in Eq.~(\ref{eq:MLEELBO}), we need the following losses to constrain CVAE latent variables sampled from Eq.~(\ref{eq:regnet_c}).  First, $\z_c^i$ and $\z_i$ should be as dissimilar as possible.  Moreover, to balance the effect of $\z_c$ and $\z_i$, we force their length to be nearly the same and yield the following loss:  
%\begin{equation*}
\begin{align}
    %\label{eq:zloss}
    L_{\z^i} = \frac{\z_c^{iT}\z_{i}}{\|\z_c^i\|\|\z_{i}\|}+\left(\|\z_c^i\|-\|\z_i\|\right)^2.
\end{align}
%\end{equation*}

%\subsubsection{Constraints on CVAE Latent Variables}  

%Figure~\ref{fig:VAE} illustrates the generation procedure of the CVAE latent variables: at the $i$-th stage, the common semantic latent variable $\z_c^i$ captures all invariant information across different domains and differs as much as possible from the specific latent variable $\z_i$.  Moreover, because $\z_c^i$ and $\z_i$ will be coupled with $\h^{i-1}$ as encoder to generate responses, we then require their length to be nearly the same to balance the impact of $\z_c^i$ and $\z_i$ in the generation procedure.  These yield the following loss:

Second, we expect that $\z_c$ changes little across two consecutive stages and enforce it by minimizing the {\em Fr\'echet Inception Distance} (FID)~\cite{DBLP:conf/nips/HeuselRUNH17}: 
%\begin{equation*}   %
\begin{align}
\label{eq:z_c_loss}
L_{\z_c^i} = \mbox{FID}(\z_c^{i-1}, \z_c^{i}). 
    %L_{\z_c}^i = %{\cal D}(\z_c^{i-1}, \z_c^i).
    %\|\z_c^i - \z_c^{i-1}\|^2.
\end{align}
%\end{equation*}
This loss is also equivalent to minimizing the Wasserstein-2 distance on two isotropic Gaussian distributions, i.e., the sum of the difference of mean and standard deviation on two Gaussian distributions.
%where $\bpsi_c^{i}$ is defined in Eq.~(\ref{eq:posteriornet_c}).
%More specifically, we minimize the above loss by setting the first two  moments of the consecutive stages attained from Eq.~(\ref{eq:posteriornet}) as close as possible. 

Third, to guarantee encoding meaningful information, we follow the idea of~\cite{DBLP:conf/acl/ZhaoZE17} to enforce the bag-of-word (BOW) loss on $\z_c^i$:
%\begin{equation*} %  
\begin{align}
\label{eq:bow_loss}
    L_{\z_c^i}^{bow} = E_{\z_c^i\sim q_{\bpsi, c}^i}[\log p(\y_{bow}|\h^i, \z_c^i)],
\end{align}
%\end{equation*}
where $\y_{bow}$ are the words in response $\y$ without order, and $p(\y_{bow}|\h^i, \z_c^i)$ is obtained by a single layer fully-connected network $\h^b = \MLP_b([\h^i; \z_c^i])$.

%where $\z_c^{i-1}$ denotes the common features distribution status from the previous stage. 

%{\bf Restriction on the attribute effect.} , we hope $\z_i$ can successfully classify the corresponding attribute and yield the following cross entropy loss
Fourth, the cross entropy loss is placed to guarantee the effect of the fed attribute:
\begin{align}  \label{eq:cls_loss}
    %\begin{aligned}
  L_{cls}^i = -\y_{l_i}\log\left({\MLP_i(\z_{i})}\right). 
    %\end{aligned}
\end{align}
%where $\ty_{pred}^i = \MLP_i(\z_{i})$ denotes the classification prediction score from a fully-connected neural network at the $i$-th stage.  %$L_{cls}^i$ defines the cross entropy between the prediction score and the one-hot representation associated with the label $l_i$.  $\MLP_i$
%to identify different styles.

%Intuitively, we wish the style feature $\z_{i}$ should mainly contain the style feature.  For a better representation, we design a classification task to force style features to have different shapes class-wise. 
%For style feature $z_{i}$ in $i$ stage, we use a single layer classifier to force $z_{i}$ to learn to identify different styles.
%The loss could be described as:

\subsubsection{Multi-stage Training}
We train PHED progressively: at the first stage, we estimate the model parameters by minimizing the following loss:
\begin{align}%\label{eq:lstage1}
%\begin{align*}
    %&
 \!\! \!\!  L_{stage}^{1}\! =\! \lambda(L_{KL}^{c, 1} + L_{KL}^1) + L_{M}^1 %\\    &\qquad\qquad 
    + L_{\z^1} + L_{\z_c^1}^{bow} + L_{cls}^1,
%\end{align*}
%\nonumber     
\end{align}
where $\lambda$ is gradually increased from 0 to 1 via the annealing technique~\cite{DBLP:conf/acl/HuangKGx18} because $L_{M}^1$, $L_{\z_c^1}^{bow}$, and $L_{cls}^1$ are cross entropy losses with nearly the same scale while the effect of $L_{\z^1}$ is small as observed.

 %the following loss:  

%For the stages from After well trained parameters within stage 1, we freeze parameters related to style label 1 (represented as the blue area in fig. \ref{fig:PHED_desc}).
%Then progressively train stage2 with loss $L_stage_{2}$ as it shows in eq. \ref{eq:lstage2}.(also $\z_c^i$ and  $\z_i$) 
%\begin{equation}  \label{eq:lstage2}
Next, at the $i$-th stage ($i\!\ge\!2$), we freeze previously learned parameters and seek new parameters by minimizing
\begin{equation}
    L_{stage}^{i}\! =\! \lambda(L_{KL}^{c, i}\! +\! L_{KL}^i)\! +\! L_{M}^i %\\     &\qquad\qquad 
    \!+\! L_{\z^i}\! +\! L_{\z_c^i}^{bow}\! +\! L_{cls}^i\! +\! L_{\z_c^i},
%\end{equation*}
\end{equation}
where the loss $L_{\z_c^i}$ is specially included to guarantee the smoothness of the change of $\z_c^i$.  The above minimization procedure continues until $i$ reaches $K$.

After training PHED, given a message $\x$, we can then generate each type of responses with the associated controlled attribute at each stage.  That is, we sample $\z_c^i$ and $\z_i$ from Eq.~(\ref{eq:priornet_i}) and concatenate them with $\h^i$ to construct the input of Transformer at each stage, i.e., $\th^1=[\h^0; \z_c^1; \z_1]$ for $E_{dec}^{d_1}$ as in Eq.~(\ref{eq:h1}) and $\th^i=[\h^{i-1}; \z_i]$ for $E_{dec}^{d_i}$ as in Eq.~(\ref{eq:h_i}), where $i=2, \ldots, K$.  Let $E_{dec, t}^k$ be the $k$-th stage decoder at the $t$-th step, we can generate the response by %as~\cite{DBLP:conf/nips/VaswaniSPUJGKP17}:% by applying the masking mechanism as:  
\begin{align}%\nonumber
%{\bf C}_t &= \tanh([\tilde{\z}_K, \h^{K}, E_{dec}^{d_K}(\y_{0, \ldots, t-1})]) \\\nonumber
%{\bf O}_t &= Dec(\h^i, ), %\MLP_o({\bf C}_t)\\\nonumber
%\P_t &= {\rm softmax}(E_{dec, t}^kW_o)\\\label{eq:output}
\y^k_t & \sim {\rm softmax}(E_{dec, t}^kW_o),  
\end{align}
where $W_o \in\R^{H\times|V|}$ is the parameter shared with the embedding layers and $|V|$ is the vocabulary size. 
 
\begin{table}[!htp]
\small
\centering
\begin{tabular}{l@{~~}rcccr}
%{lrcccr@{~~}}
%\hline & \multirow{2}{*}{\#. Pairs} & \multirow{2}{*}{min.} & avg.  & \multirow{2}{*}{\#. Char.} \\ 
\hline
%& \multicolumn{1}{c}{\#. Pairs} & \multicolumn{1}{c}{min} & max & avg. & \multicolumn{1}{c}{\#. Char.}  \\ 
& \multicolumn{1}{c}{\#. Pairs} & min. & max. & avg. & \#. Char.\\
\hline 
%\multirow{2}{*}{\bf Train} & post 
{\bf Train}$_{\rm m}$
& \multirow{2}{*}{3,542,103} & {13} &59& {38.0\small{$\pm$10.6}} & {10,261}\\%\hline& resp.
{\bf Train}$_{\rm r}$ & &{5}&{30} &{28.9 \small{$\pm$10.5}} & {6,603}\\\hline%\multirow{2}{*}{\bf Valid.} & post 
{\bf Valid.}$_{\rm m}$ & \multirow{2}{*}{196,783}& {13}&59 & {38.1\small{$\pm$10.6}} & {5,693} \\%\hline & resp. 
{\bf Valid.}$_{\rm r}$ &  &{9}&30 & {28.9\small{$\pm$10.6}} & {6,064}\\\hline%\multirow{2}{*}{\bf Test} & post 
{\bf Test}$_{\rm m}$ & \multirow{2}{*}{196,783}& {13} &59 &{38.1\small{$\pm$10.6}} & {5,714}\\ %& resp. 
{\bf Test}$_{\rm r}$
&   &{7}&30 & {28.9\small{$\pm$10.5}} & {6,126} \\
\hline
\end{tabular}
\begin{tabular}{l|c|c|c|c|c|c}
    \hline
     & Type & {Train } & {Test } & Type & {Train } & {Test }\\\hline
\multirow{3}{*}{Emo.}%\rotatebox{90}{emotions}} 
& ${\rm A}$ & 4.2  & 4.2  %\\
& ${\rm D}$ & 23.0  & 23.3  \\
& ${\rm H}$ & 5.1  & 5.2  %\\
& ${\rm L}$ & 22.0  & 22.1  \\
& ${\rm S}$ & 10.8  & 10.8
& ${\rm O}$ & 34.8  & 34.6  %\\
  \\\hline
\multirow{2}{*}{Tone} % \shortstack{sentence\\functions}}
 & ${\rm D}$ & 61.5 & 61.6  %\\
 & ${\rm I}$ & 18.0  & 17.9  \\
& ${\rm M}$ & 20.6  & 20.5  & & &  \\\hline 
\multirow{1}{*}{Len.} 
& ${\rm L}$ & 58.0  & 58.1  %\\
& ${\rm S}$ & 42.0 & 41.9  \\\hline
    \end{tabular}
\caption{\label{tb:weibo_stat} Statistics of the data.}
\end{table}
\begin{CJK*}{UTF8}{gbsn}
\section{Experiments}
%In this section, we first list our experimental data with the statistics.  After that, we depict the comparison methods and their implementation settings.  We then present the evaluation metrics and detail our experimental results.  Finally, we provide several examples to illustrate how we can generate vivid responses with the controlled attributes. ~\footnote{\href{http://weibo.com}{http://weibo.com}}
%We conduct experiments on a large-scale dataset to demonstrate the advantages of PHED.  (valid.)
We conduct experiments to address the following questions: (1) What is the performance of PHED in both automatic and human evaluations? (2) What is the effect of the losses in PHED? (3) What are the generation results? 

% , a Chinese social platform
\subsection{Data}
The data is the short-text conversation dataset  (\textbf{STC})~\cite{DBLP:conf/acl/ShangLL15}, collected from Sina Weibo, a Chinese social platform.  After setting the maximum number of characters in a response to 30, we obtain around 3.9 million dialog pairs and split them into the set of training, validation, and test with the ratio of 90\%, 5\%, and 5\%, respectively.  We pick three independent aspects of attributes, Emotion ({\bf Emo.}), {\bf Tone}, and length ({\bf Len.}).  The emotion aspect consists of six categories: angry (${\rm A}$), disgust (${\rm D}$), happy (${\rm H}$), like (${\rm L}$), sad (${\rm S}$), and others (${\rm O}$).  The tone aspect considers three types: declarative (${\rm D}$), interrogative (${\rm I}$), and imperative (${\rm M}$).  The emotion classifier and the tone classifier is trained as in~\cite{DBLP:conf/aaai/ZhouHZZL18,DBLP:conf/acl/HuangKGx18}.  Based on the typical length generated by the baselines, we set the length of a response as long, denoted by ${\rm L}$, when the number of characters is greater than 12 and others as short, denoted by ${\rm S}$.  Table~\ref{tb:weibo_stat} reports the data statistics.  %More details about the datasets are described in the Appendix. 

\begin{table*}
%Performance of relevance and diversity under the automatic evaluation and   where . 
%$^\ast$ For MMI-bidi model, the metric is calculated with $\lambda = 0.2$. $^\star$ For SC-S2S model, the metric is calculated with specificity $s=0.5$. %$^\diamond$ %For PHED model, we use the controlling labels set $<$EMOT$\_$HAPPY$>$, $<$FUNC$\_$DEC$>$, $<$LEN$\_$SHORT$>$.
\centering
\begin{tabular}{|l|c|c|c|c|c|c|c|c|c|c|}%{|@{~}l@{}|@{}c@{}|@{}c@{}|@{}c@{}|@{}c@{}|@{}c@{~}|@{~}c@{~}|@{}c@{}|@{~}c@{~}|@{~}c@{~}|@{}c@{}|}
\hline \multirow{2}{*}{{Method}} & \multicolumn{4}{|@{}c@{}|}{\bf Relevance} & \multicolumn{2}{|@{}c@{}|}{\bf Diversity (\%)}  & \multicolumn{3}{|@{}c@{~}|}{\bf Human Evaluation}& \multicolumn{1}{|@{}c@{}|}{\bf Len.}\\\cline{2-11} 
 & BLEU-1 & BLEU-2 & BLEU-3 & BLEU-4 & Dist.\,1 & Dist.\,2 & Quality & Good & Accept& \multicolumn{1}{|c|}{\bf Avg.$\pm$ \bf Std.}\\\hline
MMI-bidi%$^\ast$ 
& 10.26 & 6.03 & 4.18 & 3.20 & 0.15 & 2.61 &1.56\small{$\pm$0.57}&16\%&48\%& 11.3\small{$\pm$1.9}\\
SC-S2S%$^\star$  
& 9.63 &  5.92 & 3.87 & 2.95 & \underline{ 0.17} &  2.47  & 1.50\small{$\pm$0.55}&18\%&48\%&10.1\small{$\pm$2.2}\\ 
DCVAE & 11.72 & 6.97 & 4.82 & {3.88} & \bf 0.18 & 3.07 &1.73\small{$\pm$0.94}&24\%&54\%& 9.1\small{$\pm$2.6}\\ %\hline
T2T & 17.11 & 9.26 & 6.16 & 4.63 & 0.13 & 2.81 & 1.83\small{$\pm$0.90} & 24\% & 56\% &  12.0$\pm$1.9\\\hline  
{PHED}$_{\rm H}$ & { 16.75} & { 9.03} & { 6.00} & 4.43 & 0.14 & { 3.72}  &2.03\small{$\pm$0.90}&{\bf 42\%}&{66\%}&12.9\small{$\pm$3.1}\\
{PHED}$_{\rm S}$ & { 17.69} & { \underline{9.69}} & { \underline{ 6.55}} & \bf 4.95 & 0.13 & { \underline{4.26}}  &1.98\small{$\pm$0.89}&{ 39\%}&{58\%}&14.4\small{$\pm$3.8}\\
{PHED}$_{\rm HD}$ & { 17.04} & { 9.17} & { 6.08} & 4.51 & 0.14 & { 3.74}  &\underline{2.08\small{$\pm$0.79}}&{{40\%}}&{\underline{70\%}}&13.3\small{$\pm$3.1}\\
{PHED}$_{\rm HI}$ & { 15.61} & { 8.11} & { 5.21} & 3.78 & 0.12 & { 2.96}  &2.01\small{$\pm$0.82}&{33\%}&{54\%}&13.6\small{$\pm$3.2}\\
{PHED}$_{\rm SD}$ & { 18.28} & {\bf 9.82} & {\bf 6.57} & \underline{4.94} & 0.13 & { \underline{4.26}}  &2.02\small{$\pm$0.86}&{39\%}&{56\%}&14.6\small{$\pm$3.8}\\
{PHED}$_{\rm SI}$ & { 14.43} & { 8.02} & { 5.44} &4.14 & 0.13 & { 3.64}  &1.96\small{$\pm$0.88}&{30\%}&{61\%}&13.6\small{$\pm$3.5}\\
{PHED}$_{\rm HDS}$ & { 14.26} & { 7.56} & { 4.95} & 3.64 & 0.15 & { 3.52}  &2.01\small{$\pm$0.96}&{\underline{41\%}}&{65\%}&11.2\small{$\pm$1.5}\\
{PHED}$_{\rm HDL}$ & { \bf 18.69} & { \bf 9.82} & { { 6.37}} & {{4.65}} & 0.11 & { 3.62}  &1.95\small{$\pm$0.95}&{35\%}&64\%&15.6\small{$\pm$3.1}\\
{PHED}$_{\rm HIS}$ & { 12.48} & { 6.47} & { 4.12} & 2.99 & 0.14 & { 2.92} &{\bf 2.10\small{$\pm$0.82}} &31\%&{\bf 71\%}&11.3\small{$\pm$2.1}\\
{PHED}$_{\rm HIL}$ & { 16.67} & { 8.53} & { 5.47} & 3.88 & 0.11 & { 2.90}  &1.93\small{$\pm$0.86}&25\%&59\%&16.0\small{$\pm$3.0}\\
{PHED}$_{\rm SDS}$ & {14.47} & { 8.07} & { 5.41} & 4.21 & 0.16 & { \bf 4.61} &1.87\small{$\pm$1.06}&35\%&54\%&11.3\small{$\pm$1.9} \\
{PHED}$_{\rm SDL}$ & {\underline{18.06}} & {9.53} & { 6.28} & {4.65} & 0.15& { {4.06}}  &1.86\small{$\pm$0.99}&31\%&54\%&16.8\small{$\pm$3.3}\\
{PHED}$_{\rm SIS}$ & { 11.86} & {6.67} & {4.67} & 3.59 & 0.13& { 3.83} &1.81\small{$\pm$0.97}&28\%&53\% &11.2\small{$\pm$1.8}\\
{PHED}$_{\rm SIL}$ & { 15.88} & { 8.57} & { 5.69} &  4.52 & 0.11 & { 3.53} &1.77\small{$\pm$0.89} &22\%&47\%&16.4\small{$\pm$3.2}\\
\hline
\end{tabular}
\caption{\label{tb:stc_eval}
Evaluation results of all compared methods, where PHED reports 14 cases by selecting two typical types in each aspect of attributes.  The best two results are highlighted by bold and an underline, respectively.
%we report PHED on 14 attribute settings for response generation by only selecting two categories in each aspect of attribute.   The good (accept) ratio is the percentage of responses with 3-point (at least 2-point).  The best two results are highlighted by bold and an underline, respectively.
}
\end{table*}

\subsection{Methods and Implementation} %  and Implementation Details%(3) {\bf CT}: the conditional training method that feeds the embedding of pre-defined response attributes to the decoder of a sequence-to-sequence model~\cite{DBLP:conf/naacl/SeeRKW19}; 
%\label{sec:methods}
%\noindent{\bf Comparison Methods and Implementation}
We compare our PHED with the following strong baselines: (1) {\bf MMI-bidi}~\cite{DBLP:conf/naacl/LiGBGD16}: a popular Seq2Seq LSTM model applying the maximum mutual information objective to re-rank responses from beam search; (2) {\bf SC-S2S}~\cite{DBLP:conf/acl/ChengXGLZF18}: an Seq2Seq LSTM model with the specificity controlling; (3) {\bf DCVAE}~\cite{DBLP:conf/emnlp/GaoBLLZS19}: a newly proposed Seq2Seq LSTM model with CVAE to generate responses through two-stage sampling.
(4) {\bf T2T}~\cite{DBLP:conf/nips/VaswaniSPUJGKP17}: a Transformer-based baseline, which is also the base of PHED without controllable attributes.
%\,\footnote{\href{https://github.com/fxsjy/jieba}{https://github.com/fxsjy/jieba}}.  Specificity-Controlled-Generation
%We apply the default settings in the original papers for the baseline models.  {MMI-bidi} is reimplemented for the Chinese dialog task with the Chinese-character-level tokenization as PHED.  {SC-S2S} is re-implemented from the official open-source repository\,\footnote{\href{https://github.com/daqingchong/Specificity-Controlled-Generation}{https://github.com/daqingchong/}} by applying Jieba  segmenter.  DCVAE is directly applied the source code\,\footnote{\href{https://ai.tencent.com/ailab/nlp/dialogue}{https://ai.tencent.com/ailab/nlp/dialogue}} with a provided segmenter to tokenize the Chinese words.  Finally, SC-S2S constructs a vocabulary dictionary with 40,000 most frequent Chinese words and DCVAE constructs a vocabulary dictionary with 50,000 most frequent Chinese words and characters. %, covering 99.8\% of total words and characters.  covering 96.8\% of total words 
%说明我们如何实现的baseline

%\subsection{Implementation Details} % \,\footnote{\href{https://pytorch.org}{https://pytorch.org}}as~\cite{DBLP:conf/acl/LeSCH19}
%\label{sec:subimplementationd}DBLP:conf/nlpcc/TanSSL19,
Our implementation is in PyTorch\,\footnote{\tiny{\url{https://www.dropbox.com/s/1376kmhvuaxqe5h/PHED.zip?dl=0}}}.  We apply the default setup for the baselines with the best tuned parameters on the highest BLEU score.  Following~\cite{DBLP:conf/acl/ChengXGLZF18}, SC-S2S applies Jieba to construct a vocabulary dictionary in 40,000 most frequent Chinese words.  The same as~\cite{DBLP:conf/emnlp/GaoBLLZS19}, DCVAE constructs a vocabulary dictionary with 50,000 most frequent Chinese words and characters.  MMI-bidi and T2T apply the same Chinese characters as the input to attain a dictionary with 9,500 Chinese characters out of total 10,549 characters extracted from the dataset, where the low-frequented 1,049 characters are denoted by $<$UNK$>$.  Moreover, three additional special tokens, $<$EOS$>$, $<$BOS$>$, and $<$PAD$>$ are introduced to denote the start and the end of a sentence, and the padding character, respectively.  By this setting, we obtain a much smaller size of the dictionary in T2T and PHED, yielding a lighter effort to learn precise representations for the input tokens.  For each Transformer block, we set the number of self-attention heads to 8 and the hidden size ($H$) to 512.  T2T consists of 6 Transformer layers in both the encoder and the decoder, pre-trained on 40 million message-response pairs.  PHED applies CVAE on top of T2T.  The size of CVAE latent variables is set to 128, i.e., $d_{\z_c}=d_{\z_i}=128$.  We stack two more Transformer layers for each aspect of attributes in the order of emotion, tone, and length and yield a total of 12 Transformer layers in PHED.  The order of fed attributes is not optimized but follows the proved humans' effective learning procedure~\cite{journals/PS/SpieringA08}, from difficult tasks to easy ones.  Except for the parameters in the initial Transformer block, the remaining parameters are initialized by the \textbf{Xavier} method and trained by \textbf{ADAM} with the learning rate 0.0001 and the batch size of 32.  In the inference, we set the beam search size to 5.  Under the above settings, we train PHED 10 epochs at each stage on a Tesla V100 GPU and cost about 51 hours.

%three more Transformer blocks with two s each and yield a total of 12 encoder-decoder Transformer s.  Our PHED consists of 12 encoder-decoder s when applying for all attributes. ~\cite{DBLP:journals/jmlr/GlorotB10} ~\cite{DBLP:journals/corr/KingmaB14}  following~\cite{DBLP:conf/emnlp/GaoBLLZS19},

%The number of s in the initial Transformer blocks~\cite{DBLP:conf/naacl/DevlinCLT19} for both the encoder and the decoder is 6, trained by a pre-trained language model on 40 million message-response pairs, following the standard procedure in~\cite{DBLP:conf/nlpcc/TanSSL19,DBLP:conf/acl/LeSCH19}. 

% Delete after complete~\cite{DBLP:conf/acl/PapineniRWZ02}

\subsection{Evaluation Metrics} %  and Results
%\noindent{\bf Evaluation Metrics} 
We evaluate the models by: (1) {\bf BLEU}: BLEU-$n$ measures the average $n$-gram precision on a set of reference sentences.  As DCVAE~\cite{DBLP:conf/emnlp/GaoBLLZS19}, we set $n=1, 2, 3, 4$. (2) {\bf Dist.\,1 \& Dist.\,2}~\cite{DBLP:conf/naacl/LiGBGD16}: the ratios of distinct unigrams and bigrams in the generated responses to the total generated unigrams and bigrams, measuring the diversity of the responses.  For a fair comparison, all metrics are evaluated by the Chinese-character-level tokenization.  (3) {\bf Human evaluation}: Three expert labelers were recruited to evaluate the generated responses for 300 randomly selected posts based on the following 4-point criteria: 1) +3: the response is not only semantically relevant and grammatically correct, but also informative and interesting; 2) +2: the response is grammatically correct and can be used as a response to the utterance, but is too general (e.g., ``OK''); %\begin{CJK*}{UTF8}{gbsn}(“真好啊”)\end{CJK*} ``Great''; 
3) +1: the response is grammatically correct, but semantically irrelevant; 4) +0: the response contains mistakes (e.g., grammatical errors or $<$UNK$>$).  Though it is different from the 3-point criteria in~\cite{DBLP:conf/acl/ChengXGLZF18,DBLP:conf/emnlp/GaoBLLZS19}, the 4-point criteria allows us to further distinguish meaningful responses from general and irrelevant responses.  The values of the Fleiss’ Kappa~\cite{journals/ESM/Fleiss73} are great than 0.3 in all cases, which indicate the inter-rater consistency among three labelers.

% metrics are firstly given in  to compare our method and baseline methods. in MMI-bidi, $\lambda=0.2$ and in SC-S2S, the specificity $s=0.5$, which%a letter in the subscription denotes the corresponding attribute at the corresponding stage.  The first letter in the subscription, ${\rm H}$ and ${\rm S}$, denotes the emotion to {\em Happy} and {\em Sad}, respectively.  The second letter in the subscription, ${\rm D}$ and ${\rm I}$, denotes the type of the sentence function to {\em Declarative} and {\em Interrogative}, respectively.  The third letter in the subscription, ${\rm S}$ and ${\rm L}$, denotes the response length to {\em Short}  and {\em Long}, respectively.  Here, ${\rm H}$ and ${\rm S}$ are selected because their corresponding training data is the minimum and median size in emotion domain.  ${\rm D}$ and ${\rm I}$  correspond to the maximum and median training data in the sentence function domain.  As defined in Sec.~\ref{sec:ex_data}, {PHED}$_{\rm HIS}$ denotes PHED generates a {\em Short} responses with the {\em Happy} emotion and the {\em Interrogative} sentence function.  

% i.e., ${\rm S}$ in the third char of the subscript in PHED, i.e., ${\rm L}$, in the third char of the subscript in PHED, %, \blue{except BLEU-4 for {PHED}$_{\rm HDL}$}. 
%The automatic evaluation results show that  i.e. PHED$_{\rm HDS}$, PHED$_{\rm HIS}$, PHED$_{\rm SDS}$, and PHED$_{\rm SIS}$,

\subsection{Experimental Results} %  and Analysis where the parameters in the baselines are tuned to yield the highest BLEU score.  To show the effectiveness of , where the first character indicates the aspect of emotions, the second character indicates the aspect of sentence functions, and the third character indicates the required response length.  According to the distribution statistics in Table~\ref{tb:labels_dist}, we select the attributes, which contain no bias to any specific aspect. and yields 14 cases
%These results reveal that PHED not only generates responses with multiple attributes, but also relevant to the posts. 
%\noindent{\bf Experimental Results} 
%Table~\ref{tb:stc_eval} reports the evaluation results, where PHED selects two typical types for each aspect according to the statistics shown in Table~\ref{tb:labels_dist}.  Results in Table~\ref{tb:stc_eval} 

%PHED performs similarly on all the combinations of attributes.  Due to the space limitation, we select two typical types for each aspect of attributes, which yield 14 cases of PHED, and report the results with those in baselines in Table~\ref{tb:stc_eval}.  The results show that  
Table~\ref{tb:stc_eval} reports 14 cases of PHED by selecting two typical types for each aspect of attributes and shows that 
\begin{compactitem}[-]
\item {\bf Relevance.}  PHED attains the best two BLEU scores, which imply the generation relevance.  Moreover, long responses can get significantly higher BLEU scores ($p<0.01$ in $t$-test) than the short responses and the LSTM-based methods.  Even short responses can attain competitive BLEU scores compared to the baselines.  
\item {\bf Diversity.}  PHED attains relatively lower scores in {Dist.\,1} because usually PHED generates longer responses than baselines, yielding a larger number of unigrams.  In terms of {Dist.\,2}, PHED attains significantly higher scores than the baselines (e.g., 4.61 vs. 3.07, around 50\% gain).  This again shows that PHED generates more diverse responses.
\item {\bf Length.} According to the results in the last column, PHED tends to generate longer responses, with more powerful expression ability than the baselines.  Even when setting to generate short responses, PHED can generate longer responses than SC-S2S and DCVAE and similar length for MMI-bidi and T2T.  When setting to generate long responses, PHED can generate 3 to 5 more characters for each response, significantly longer than the baselines.
%, we argue that the lower Distinct-1 score is reasonable.  Since we use the same Chinese-character-level tokenizer and the test set contains more than 190k utterances, the distinct unigrams set is approximately the same as the vocabulary set. \blue{ Based on statistic analysis of the average and standard deviation of generated response lengths, as shown in Tab. \ref{tb:stc_eval}, we argue that for those methods which generate longer responses, it's reasonable to have a compared lower Distinct-1 score in this experiment. } The larger total number of generated tokens lead the slightly lower Distinct-1 score in this experiment.  In that case, the dominating factor of the Distinct-1 is the total generated tokens. 
\end{compactitem}

The human evaluation results in the eighth to tenth columns of Table~\ref{tb:stc_eval} are consistent with the automatic evaluation: (1) PHED generates significantly more relevant responses, where the values of 42\%~vs.~24\% and 71\%~vs.~56\% indicate that PHED generates more good (scoring over 3-point) and acceptable (scoring over 2-point) responses. (2) DCVAE and T2T are competitive while MMI-bidi and SC-S2S attaining much lower scores than them.  Overall, PHED generates more good responses than baselines.

\begin{table}
%Performance of relevance and diversity under the automatic evaluation and   where .  on the losses
%$^\ast$ For MMI-bidi model, the metric is calculated with $\lambda = 0.2$. $^\star$ For SC-S2S model, the metric is calculated with specificity $s=0.5$. %$^\diamond$ %For PHED model, we use the controlling labels set $<$EMOT$\_$HAPPY$>$, $<$FUNC$\_$DEC$>$, $<$LEN$\_$SHORT$>$.\multirow{1}{*}{{}} 
\centering
\begin{tabular}{|@{~}l@{~}|@{~}c@{~}|@{~}c@{~}|@{~}c@{~}|@{~}c@{~}|@{~}c@{~}|@{~}c@{~}|}%{|@{~}c@{~}|@{~}c@{~}|@{~}c@{~}|@{~}c@{~}|@{~}c@{~}|@{~}c@{~}|}
\hline & BLEU & Dist. & Emo. & Tone & {Len. A.} &  {Len.} 
\\\hline
T2T & 4.63 & 2.81 & $-$ & $-$ & $-$ & 12.0\small{$\pm$1.9} \\\hline
PHED$_{\rm S}$ & 4.96 & 4.26 & 54.6 & $-$ & $-$ & 14.4\small{$\pm$3.8} 
% 17.70 & 9.69 & 6.55 & 4.96 & 0.13 & 4.26 
\\\hline
PHED$_{\rm SI}$ & 4.14 & 3.64 & 57.6 & 93.8 & $-$ & 13.6\small{$\pm$3.5}
% 14.43 & 8.03 & 5.45 & 4.14 & 0.13 & 3.64 
\\\hline
PHED$_{\rm SIS}$ & { 3.59} & {3.83} & {60.3} & 89.3 & 84.0 & 11.2\small{$\pm$1.8} \\\hline
%$-$ CVAE & 1.52 & 0.94 & {-} & - &   9.78 \\
\small{$-L_{cls}^i$}  & { 5.57} & {5.67} & {9.7} & 21.6 & 48.6 & 13.6\small{$\pm$2.8} \\
%\small{$-L_{\z_{c}^i}$} & { 3.20} & {3.69} & {61.16} & 91.40 & 86.5 & 11.0\small{$\pm$1.9}  \\ 
%\small{$-L_{\z^{i}}$} & { 3.23} & {3.97} & {62.15} & 91.36 & 85.9 & 11.1\small{$\pm$1.9} \\ 
%\small{$-L_{\z_{c}^i}^{bow}$}  & { 3.13} & {3.72} & {63.82} & 91.15 & 87.8 & 11.0\small{$\pm$1.9} \\
%\small{$-L_{\z_{c}^i}^{bow}-L_{\z_{c}^i}$}  & { 3.07} & {3.63} & {64.44} & 91.78 & 87.2 & 10.9\small{$\pm$2.0} \\
\small{$-L_{\z^{i}}\!-\!L_{cls}^i$} & { 3.19} & {3.84} & {62.4} & 90.6 & 88.2 & 10.9\small{$\pm$2.1} \\
\hline
PHED$_{\rm SIL}$ & { 4.52} & { 3.53} & {59.6} & {90.8}  & 94.9 & 16.4\small{$\pm$3.2} \\\hline
%$-$ CVAE & 2.63 & 1.02 & {-} & - &   13.53 \\
\small{$-L_{cls}^i$}  & { 5.57} & {5.68} & {9.8} & 21.6 & 51.4 & 10.9\small{$\pm$2.8} \\
%\small{$-L_{\z_{c}^i}$} & 3.97 & 3.43 & {59.85} & 92.68 & 94.4 &   16.2\small{$\pm$3.1}  \\
%\small{$-L_{\z^{i}}$}  & 3.93 & 3.56 & {61.95} & 92.52 & 94.1 &   16.3\small{$\pm$3.0}   \\ 
%\small{$-L_{\z_{c}^i}^{bow}$}  & { 3.97} & {3.41} & {59.65} & 92.33 & 94.4 & 16.2\small{$\pm$3.0} \\
%\small{$-L_{\z_{c}^i}^{bow}-L_{\z_{c}^i}$}  & { 3.83} & {3.33} & {61.69} & 92.58 & 94.4 & 16.2\small{$\pm$3.1} \\
\small{$-L_{\z^{i}}\!-\!L_{cls}^i$} & {3.91} & 3.45 & {61.8} & 91.4 & 96.4 & 16.5\small{$\pm$3.0}  \\
\hline
\end{tabular}
\caption{\label{tb:ablation_study}
Results on ablation study.}
\end{table}

\subsection{Ablation Study}
%\noindent{\bf Ablation Study}
We conduct ablation studies on PHED.  The test shows that PHED produces similar results on all the combinations of attributes.  In Table~\ref{tb:ablation_study}, we only report the scores of BLEU-4, Dist.\,2, the accuracy of emotion (Emo.), tone (Tone), and the length prediction (Len.\,A.), and the average length (Len.) in five models: T2T, PHED$_{\rm S}$, PHED$_{\rm SI}$,  PHED$_{\rm SIS}$ and PHED$_{\rm SIL}$; see more results in the Appendix. 

The results show that: (1) The Dist.~2 scores in PHED on different fed attributes are all higher than that in T2T.  This means that PHED attains more diverse responses than T2T.  By examining more details, we can observe that the emotion accuracy increases slightly after adding more other attributes.  The tone accuracy is around 90\% while the length accuracy is at least 84\%. (2) By removing the losses related to $\z_c$ in PHED, we obtain similar BLEU-4 and Dist.~2 scores to PHED with all restricted losses but attain slightly higher accuracy on all three aspects of attributes.  The results imply that by removing $\z_c$, we can promote the effect of individual attributes and yield better attribute accuracy. (3) By examining the results of only removing classification loss (i.e., $-L^i_{cls}$), we observe that the corresponding attribute's accuracy slashes largely and becomes normal when removing both $L_{\z^i}$ and $L^i_{cls}$.  The observation means that $L^i_{cls}$ plays an essential role in controlling the attributes when PHED needs to satisfy other minimization requirements.

\begin{figure}[hpbt]
\small
\centering%\footnotesize
\begin{tabular}{@{}l@{}p{7cm}@{}} %
\hline
{\multirow{1}{*}{Post}} & 全场比赛结束。乌迪内斯2-1米兰。~ The match is over.  Udinese wins Milan 2-1. \\ \hdashline
MMI-bidi &  哈哈，真是一场精彩的比赛~  Haha, what a wonderful game\\
SC-S2S      &  比赛真好看，真好看~  The game is awesome, really good\\
DCVAE      &  米兰的比赛结束了吗？~ Is Milan's match over?\\ \hdashline

T2T & 好可爱啊！我也想要。~ So cute! I want too.  \\\hdashline 
PHED$_{{\rm S}}$ & 阿莱格里$\overleftrightarrow{\mbox{真心伤不起}}$啊。 Allegri $\overleftrightarrow{\mbox{really hurt}}$. \\\hdashline
%PHED$_{\rm \blue{\bf H}\underline{D}}$ & \underline{{\blue{\bf 恭喜}米兰}，终于可以看到这场比赛了} &\underline{\blue{\bf Congratulations} to Milan, finally, we can see this match}\\ % }\\&& \underline{
%PHED$_{\rm \blue{\bf H}\underline{I}}$ & \textbf{\blue{ 哈哈}}，\underline{米兰这是要逆天吗？} &  \textbf{\blue{Haha}}, \underline{is Milan going to offence God's will?} \\
PHED$_{\rm SD}$ & \underline{{我的心脏$\overleftrightarrow{\mbox{受不了啊}}$。}} \underline{My heart $\overleftrightarrow{\mbox{can't stand}}$ it .}\\ 
PHED$_{\rm SI}$ &  不知道\underline{为什么} 我觉得这场比赛的时候还是$\overleftrightarrow{\mbox{很难看}}$\\&
I don't know \underline{why} I think this game is still $\overleftrightarrow{\mbox{ugly}}$ \\\hdashline
PHED$_{\rm SDS}$ & \underline{我的心脏$\overleftrightarrow{\mbox{受不了}}$阿莱格里!} \\
& \underline{My heart $\overleftrightarrow{\mbox{cannot stand}}$ Allegri!}\\
PHED$_{\rm SDL}$ &  \underline{我的心脏$\overleftrightarrow{\mbox{受不了}}$啊，看到这么多人想$\overleftrightarrow{\mbox{骂}}$你！}\\& \underline{My heart $\overleftrightarrow{\mbox{can't bear}}$ to see so many people want to $\overleftrightarrow{\mbox{scold}}$} \underline{you!}  \\
PHED$_{\rm SIS}$ &  $\overleftrightarrow{\mbox{\underline{为什么}没有}}$阿森纳\underline{呢？}  $\overleftrightarrow{\mbox{\underline{Why} not}}$ Arsenal\underline{?} \\
PHED$_{\rm SIL}$ &  不知道\underline{为什么}，看到这个我$\overleftrightarrow{\mbox{心里很难受}}$。\\& I don't know \underline{why}, I $\overleftrightarrow{\mbox{feel bad}}$ to see this. \\
\hline
\end{tabular}
\caption{Responses generated by the compared methods.  In PHED, the words with $\leftrightarrow$ on top and the {\underline{words in underline}} of the generated responses indicate a high specificity to the first attribute and the second attribute, respectively.%We show the responses generated by PHED under the same combination of {\em Happy} and {\em Sad} emotions, {\em Declarative} and {\em Interrogative} sentence function, and both {\em Long} and {\em Short} response lengths as in Table~\ref{tb:stc_eval}.  
\label{fig:ex_all}}
\end{figure}
%\if 0
\subsection{Case Study}
%We present two generated cases in the Appendix\,\footnote{\tiny or in   \url{https://www.dropbox.com/s/wi51lczjnmxboth/phed_app.pdf?dl=0}}.
Figure~\ref{fig:ex_all} illustrates a complete examination on the compared methods in Table~\ref{tb:stc_eval}.  Our PHED clearly generates responses with the specific attributes progressively, including not only the corresponding emotion aspect, but also the exact tone and the length.  For example, in the Happy emotion, PHED frequently generates ``Congratulations" and ``Haha".  While in the Interrogative tone, it generates the related words, e.g., ``What" or ``Why".  Moreover, by changing the required response length from short to long, more characters can be produced.  For example, in Post\#1, PHED$_{\rm SDL}$ generates similar words in the beginning to PHED$_{\rm SDS}$ but produces one more sentence to enrich the expression than PHED$_{\rm SDS}$ .  By examining all responses generated by PHED, the fidelity to the attribute(s) is clearly confirmed.  In terms of the responses generated by MMI-bidi, SC-S2S, and DCVAE, they are usually shorter and blank.  Responses generated by T2T are a little fluctuation and cannot deliver any attribute effect.  More examples can be shown in the Appendix.%\,\footnote{\tiny or in   \url{https://www.dropbox.com/s/wi51lczjnmxboth/phed_app.pdf?dl=0}}.
%\noindent{\bf Case Study} 
%Figure~\ref{fig:ex_all} delivers a complete examination on the example in Fig.~\ref{fig:ex_illustration} for the compared methods in Table~\ref{tb:stc_eval}.  
%We present two generated cases in the Appendix\,\footnote{\tiny\url{https://www.dropbox.com/s/wi51lczjnmxboth/phed_app.pdf?dl=0}}.  PHED clearly generate responses with the specific attributes progressively, including not only the corresponding emotion aspect, but also the exact tone.  For example, in the Happy emotion, PHED frequently generates ``Congratulations" and ``Haha".  While in the Interrogative tone, it generates the related words, e.g., ``What" or ``Why".  Moreover, by changing the required response length from short to long, more characters can be produced.  For example, PHED$_{\rm SDL}$ generates similar words in the beginning to PHED$_{\rm SDS}$, but produces one more sentence to enrich the expression than PHED$_{\rm SDS}$ in Post\#1.  Similar results are shown in all compared cases in both Post~\#1 and Post~\#2.  
%\fi 
%\end{CJK*}
\end{CJK*}

\section{Conclusion}
We propose Progressively trained Hierarchical Encoder-Decoder to generate responses with multiple controllable attributes.  By incorporating CVAE into Transformer, we represent the controlled attributes by a joint latent variable and further specific latent variables, where the CVAE latent variables are then coupled with the encoding information to generate responses.  The model is then effectively trained progressively by maximizing the evidence lower bound while minimizing several subtly designed losses.  Empirical results with both automatic and human evaluations demonstrate that PHED significantly outperforms the state-of-the-art neural generation models and is able to generate more diverse responses.  

Several challenging but interesting directions will be considered in the future.  First, we only exploit three aspects of attributes in one order of generation.  It is practicable and useful to further take into account other aspects and other orders.  Second, we only apply CVAE under the Markov assumption.  It is interesting to explore more dependencies in CVAE.  Third, the current task only focuses on open-domain response generation.  It would be worthwhile to probe other tasks, e.g., text generation in the same spirit.

% to generate diverse responses.  It is interesting to explore other generation schemes, e.g., Generative Adversarial Networks. 

\bibliographystyle{named}
\bibliography{ref}

\begin{thebibliography}{}

\bibitem[\protect\citeauthoryear{Bi \bgroup \em et al.\egroup
  }{2019}]{DBLP:conf/acl/BiGLS19}
Wei Bi, Jun Gao, Xiaojiang Liu, and Shuming Shi.
\newblock Fine-grained sentence functions for short-text conversation.
\newblock In {\em {ACL}}, pages 3984--3993, 2019.

\bibitem[\protect\citeauthoryear{Fleiss and
  Cohen}{1973}]{journals/ESM/Fleiss73}
Joseph~L. Fleiss and Jacob Cohen.
\newblock The equivalence of weighted kappa and the intraclass correlation
  coefficient as measures of reliability.
\newblock {\em Educational and psychological measurement}, 33(3):613– 619,
  1973.

\bibitem[\protect\citeauthoryear{Gao \bgroup \em et al.\egroup
  }{2019}]{DBLP:conf/emnlp/GaoBLLZS19}
Jun Gao, Wei Bi, Xiaojiang Liu, Junhui Li, Guodong Zhou, and Shuming Shi.
\newblock A discrete {CVAE} for response generation on short-text conversation.
\newblock In {\em {EMNLP-IJCNLP}}, pages 1898--1908, 2019.

\bibitem[\protect\citeauthoryear{Heusel \bgroup \em et al.\egroup
  }{2017}]{DBLP:conf/nips/HeuselRUNH17}
Martin Heusel, Hubert Ramsauer, Thomas Unterthiner, Bernhard Nessler, and Sepp
  Hochreiter.
\newblock Gans trained by a two time-scale update rule converge to a local nash
  equilibrium.
\newblock In {\em NIPS}, 2017.

\bibitem[\protect\citeauthoryear{Hu \bgroup \em et al.\egroup
  }{2017}]{DBLP:conf/icml/HuYLSX17}
Zhiting Hu, Zichao Yang, Xiaodan Liang, Ruslan Salakhutdinov, and Eric~P. Xing.
\newblock Toward controlled generation of text.
\newblock In {\em {ICML}}, pages 1587--1596, 2017.

\bibitem[\protect\citeauthoryear{Huang \bgroup \em et al.\egroup
  }{2020}]{DBLP:journals/tois/HuangZG20}
Minlie Huang, Xiaoyan Zhu, and Jianfeng Gao.
\newblock Challenges in building intelligent open-domain dialog systems.
\newblock {\em {ACM} Trans. Inf. Syst.}, 38(3):21:1--21:32, 2020.

\bibitem[\protect\citeauthoryear{Jiao \bgroup \em et al.\egroup
  }{2019}]{DBLP:conf/naacl/JiaoYKL19}
Wenxiang Jiao, Haiqin Yang, Irwin King, and Michael~R. Lyu.
\newblock Higru: Hierarchical gated recurrent units for utterance-level emotion
  recognition.
\newblock In {\em {NAACL-HLT}}, pages 397--406, 2019.

\bibitem[\protect\citeauthoryear{Ke \bgroup \em et al.\egroup
  }{2018}]{DBLP:conf/acl/HuangKGx18}
Pei Ke, Jian Guan, Minlie Huang, and Xiaoyan Zhu.
\newblock Generating informative responses with controlled sentence function.
\newblock In {\em {ACL}}, pages 1499--1508, 2018.

\bibitem[\protect\citeauthoryear{Kikuchi \bgroup \em et al.\egroup
  }{2016}]{DBLP:conf/emnlp/KikuchiNSTO16}
Yuta Kikuchi, Graham Neubig, Ryohei Sasano, Hiroya Takamura, and Manabu
  Okumura.
\newblock Controlling output length in neural encoder-decoders.
\newblock In {\em {EMNLP}}, pages 1328--1338, 2016.

\bibitem[\protect\citeauthoryear{Lample \bgroup \em et al.\egroup
  }{2019}]{DBLP:conf/iclr/LampleSSDRB19}
Guillaume Lample, Sandeep Subramanian, Eric~Michael Smith, Ludovic Denoyer,
  Marc'Aurelio Ranzato, and Y{-}Lan Boureau.
\newblock Multiple-attribute text rewriting.
\newblock In {\em {ICLR}}, 2019.

\bibitem[\protect\citeauthoryear{Li \bgroup \em et al.\egroup
  }{2016}]{DBLP:conf/naacl/LiGBGD16}
Jiwei Li, Michel Galley, Chris Brockett, Jianfeng Gao, and Bill Dolan.
\newblock A diversity-promoting objective function for neural conversation
  models.
\newblock In {\em {NAACL}-{HLT}}, pages 110--119, 2016.

\bibitem[\protect\citeauthoryear{Logeswaran \bgroup \em et al.\egroup
  }{2018}]{DBLP:conf/nips/LogeswaranLB18}
Lajanugen Logeswaran, Honglak Lee, and Samy Bengio.
\newblock Content preserving text generation with attribute controls.
\newblock In {\em NeurIPS}, 2018.

\bibitem[\protect\citeauthoryear{Rashkin \bgroup \em et al.\egroup
  }{2019}]{DBLP:conf/acl/RashkinSLB19}
Hannah Rashkin, Eric~Michael Smith, Margaret Li, and Y{-}Lan Boureau.
\newblock Towards empathetic open-domain conversation models: {A} new benchmark
  and dataset.
\newblock In {\em {ACL}}, pages 5370--5381, 2019.

\bibitem[\protect\citeauthoryear{See \bgroup \em et al.\egroup
  }{2019}]{DBLP:conf/naacl/SeeRKW19}
Abigail See, Stephen Roller, Douwe Kiela, and Jason Weston.
\newblock What makes a good conversation? how controllable attributes affect
  human judgments.
\newblock In {\em {NAACL-HLT}}, pages 1702--1723, 2019.

\bibitem[\protect\citeauthoryear{Serban \bgroup \em et al.\egroup
  }{2016}]{DBLP:conf/aaai/SerbanSBCP16}
Iulian~Vlad Serban, Alessandro Sordoni, Yoshua Bengio, Aaron~C. Courville, and
  Joelle Pineau.
\newblock Building end-to-end dialogue systems using generative hierarchical
  neural network models.
\newblock In {\em {AAAI}}, 2016.

\bibitem[\protect\citeauthoryear{Serban \bgroup \em et al.\egroup
  }{2017}]{DBLP:conf/aaai/SerbanSLCPCB17}
Iulian~Vlad Serban, Alessandro Sordoni, Ryan Lowe, Laurent Charlin, Joelle
  Pineau, Aaron~C. Courville, and Yoshua Bengio.
\newblock A hierarchical latent variable encoder-decoder model for generating
  dialogues.
\newblock In {\em {AAAI}}, pages 3295--3301, 2017.

\bibitem[\protect\citeauthoryear{Shang \bgroup \em et al.\egroup
  }{2015}]{DBLP:conf/acl/ShangLL15}
Lifeng Shang, Zhengdong Lu, and Hang Li.
\newblock Neural responding machine for short-text conversation.
\newblock In {\em {ACL}}, pages 1577--1586, 2015.

\bibitem[\protect\citeauthoryear{Shao \bgroup \em et al.\egroup
  }{2019}]{DBLP:conf/emnlp/ShaoHWXZ19}
Zhihong Shao, Minlie Huang, Jiangtao Wen, Wenfei Xu, and Xiaoyan Zhu.
\newblock Long and diverse text generation with planning-based hierarchical
  variational model.
\newblock In {\em {EMNLP-IJCNLP}}, pages 3255--3266, 2019.

\bibitem[\protect\citeauthoryear{Spiering and
  Ashby}{2008}]{journals/PS/SpieringA08}
B.~J. Spiering and F.~G. Ashby.
\newblock Initial training with difficult items facilitates information
  integration, but not rule-based category learning.
\newblock {\em Psychol Sci.}, 19(11):1169–1177, 2008.

\bibitem[\protect\citeauthoryear{Vaswani \bgroup \em et al.\egroup
  }{2017}]{DBLP:conf/nips/VaswaniSPUJGKP17}
Ashish Vaswani, Noam Shazeer, Niki Parmar, Jakob Uszkoreit, Llion Jones,
  Aidan~N. Gomez, Lukasz Kaiser, and Illia Polosukhin.
\newblock Attention is all you need.
\newblock In {\em NIPS}, pages 5998--6008, 2017.

\bibitem[\protect\citeauthoryear{Wang and Wan}{2019}]{DBLP:conf/ijcai/Wang019b}
Tianming Wang and Xiaojun Wan.
\newblock {T-CVAE:} transformer-based conditioned variational autoencoder for
  story completion.
\newblock In {\em {IJCAI}}, 2019.

\bibitem[\protect\citeauthoryear{Xing \bgroup \em et al.\egroup
  }{2018}]{DBLP:conf/aaai/XingWWHZ18}
Chen Xing, Yu~Wu, Wei Wu, Yalou Huang, and Ming Zhou.
\newblock Hierarchical recurrent attention network for response generation.
\newblock In {\em {AAAI}}, pages 5610--5617, 2018.

\bibitem[\protect\citeauthoryear{Xu \bgroup \em et al.\egroup
  }{2019}]{DBLP:conf/acl/XuWTHSW19}
Can Xu, Wei Wu, Chongyang Tao, Huang Hu, Matt Schuerman, and Ying Wang.
\newblock Neural response generation with meta-words.
\newblock In {\em {ACL}}, 2019.

\bibitem[\protect\citeauthoryear{Zhang \bgroup \em et al.\egroup
  }{2018}]{DBLP:conf/acl/ChengXGLZF18}
Ruqing Zhang, Jiafeng Guo, Yixing Fan, Yanyan Lan, Jun Xu, and Xueqi Cheng.
\newblock Learning to control the specificity in neural response generation.
\newblock In {\em {ACL}}, pages 1108--1117, 2018.

\bibitem[\protect\citeauthoryear{Zhao \bgroup \em et al.\egroup
  }{2017}]{DBLP:conf/acl/ZhaoZE17}
Tiancheng Zhao, Ran Zhao, and Maxine Esk{\'{e}}nazi.
\newblock Learning discourse-level diversity for neural dialog models using
  conditional variational autoencoders.
\newblock In {\em {ACL}}, pages 654--664, 2017.

\bibitem[\protect\citeauthoryear{Zheng \bgroup \em et al.\egroup
  }{2020}]{DBLP:conf/aaai/ZhengZHM20}
Yinhe Zheng, Rongsheng Zhang, Minlie Huang, and Xiaoxi Mao.
\newblock A pre-training based personalized dialogue generation model with
  persona-sparse data.
\newblock In {\em {AAAI}}, pages 9693--9700, 2020.

\bibitem[\protect\citeauthoryear{Zhou and Wang}{2018}]{DBLP:conf/acl/WangZ18}
Xianda Zhou and William~Yang Wang.
\newblock Mojitalk: Generating emotional responses at scale.
\newblock In {\em {ACL}}, pages 1128--1137, 2018.

\bibitem[\protect\citeauthoryear{Zhou \bgroup \em et al.\egroup
  }{2018}]{DBLP:conf/aaai/ZhouHZZL18}
Hao Zhou, Minlie Huang, Tianyang Zhang, Xiaoyan Zhu, and Bing Liu.
\newblock Emotional chatting machine: Emotional conversation generation with
  internal and external memory.
\newblock In {\em AAAI}, pages 730--739, 2018.

\end{thebibliography}

\end{document}